\documentclass[conference]{IEEEtran}
\IEEEoverridecommandlockouts
\usepackage{cite}
\usepackage{amsmath,amssymb,amsfonts}
\usepackage{algorithmic}
\usepackage{graphicx}
\usepackage{textcomp}
\usepackage{xcolor}
\usepackage{authblk}
\usepackage{subcaption}
\usepackage{verbatim}
\usepackage{dsfont}
\usepackage{multirow}

\usepackage[top=0.71in, bottom=1in, left=0.71in, right=0.71in]{geometry}

\def\BibTeX{{\rm B\kern-.05em{\sc i\kern-.025em b}\kern-.08em
    T\kern-.1667em\lower.7ex\hbox{E}\kern-.125emX}}
\begin{document}

\title{\Large Energy-Efficient Wireless LLM Inference via \\Uncertainty and Importance-Aware Speculative Decoding}
\author{Jihoon Park, Seungeun Oh, and Seong-Lyun Kim

\IEEEcompsocitemizethanks{
    \IEEEcompsocthanksitem J. Park, S. Oh and S.-L. Kim are with the School of Electrical and Electronic Engineering, Yonsei University, Seoul, South Korea. (E-mail: \{jhpark, seoh, slkim\}@ramo.yonsei.ac.kr)
    }
    %\thanks{*Corresponding author: Seung-Woo Ko.}
    }

\maketitle
\begin{abstract}
To address the growing demand for on-device LLM inference in resource-constrained environments, hybrid language models (HLM) have emerged, combining lightweight local models with powerful cloud-based LLMs. Recent studies on HLM have primarily focused on improving accuracy and latency, while often overlooking communication and energy efficiency. We propose a token-level filtering mechanism for an energy-efficient importance- and uncertainty-aware HLM inference that leverages both epistemic uncertainty and attention-based importance. Our method opportunistically uploads only informative tokens, reducing LLM usage and communication costs. Experiments with \texttt{TinyLlama-1.1B} and \texttt{LLaMA-2-7B} demonstrate that our method achieves up to 87.5\% BERT Score and token throughput of 0.37 tokens/sec while saving the energy consumption by 40.7\% compared to standard HLM. Furthermore, compared to our previous U-HLM baseline, our method improves BERTScore from 85.8\% to 87.0\%, energy savings from 31.6\% to 43.6\%, and throughput from 0.36 to 0.40. This approach enables an energy-efficient and accurate deployment of LLMs in bandwidth-constrained edge environments.
\end{abstract}


\section{Introduction}

Large language models (LLMs), such as GPT and LLaMA, have achieved impressive breakthroughs across a wide range of natural language processing (NLP) tasks, including question answering, summarization, and dialogue generation.
These models exhibit strong generalization capabilities, serving as core enablers of AI services \cite{yang2024harnessing, lin2023pushing}.
To bring these capabilities closer to end-users, particularly in latency-sensitive or bandwidth-constrained environments, edge-based LLM inference has gained increasing attention. However, the computational demands of LLMs pose significant challenges for deployment in resource-constrained scenarios, particularly in edge-based applications such as mobile assistants, augmented reality (AR), and on-device copilots. 

To alleviate these constraints, hybrid language models (HLM) have emerged~\cite{hao2024hybrid}, where a relatively lightweight small local model (SLM) in edge generates candidate outputs, and a more capable cloud-based LLM performs verification or refinement. This framework, often instantiated via speculative decoding~\cite{leviathan2023fast}, allows lightweight on-device generation while preserving high accuracy through selective LLM involvement.
\begin{figure}[t]	
	\centering	
	\includegraphics[width = 0.48\textwidth]{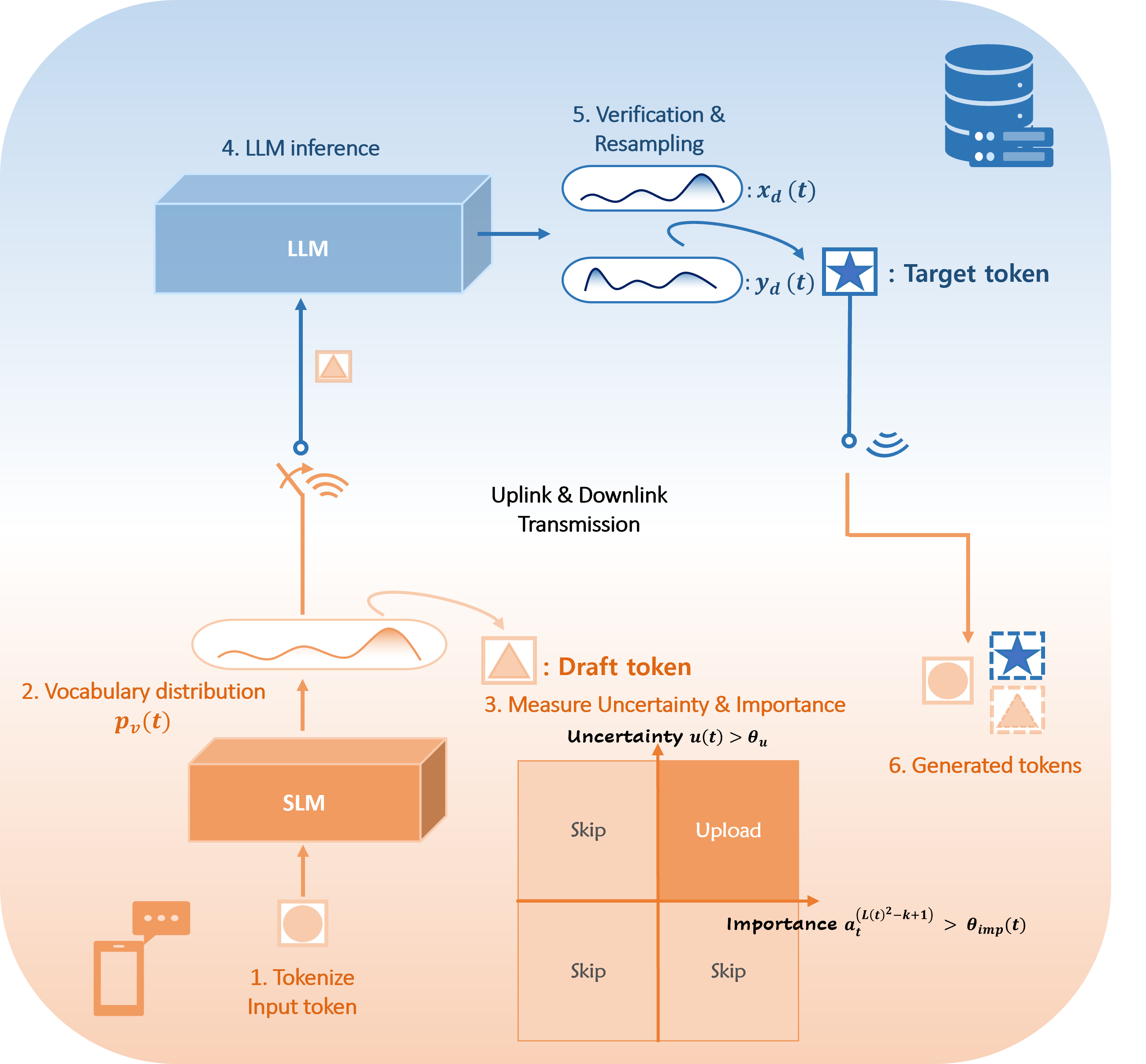}			
	\caption{Illustration of the Uncertainty and Importance-Aware Hybrid Language Model (HLM) inference architecture.}
	\label{Fig:arch}
\end{figure}
Despite the performance benefits, most existing speculative decoding approaches focus predominantly on improving latency or accuracy~\cite{zhou2023distillspec}, while overlooking key system-level constraints such as bandwidth usage and energy consumption. In many practical deployments, communication between the SLM and LLM is a major bottleneck: uploading every token's full probability distribution incurs substantial payload overhead, and excessive LLM queries further increase energy cost and latency. As LLMs are expensive to run and offloading to the cloud requires reliable uplink capacity, these issues are not just technical inconveniences, but critical barriers to scalability. Our previous works, \textit{Uncertainty-aware HLM}~\cite{oh2024uncertainty, park2025uncertaintyaware}, attempted to address these issues by selectively uploading only the tokens with high epistemic uncertainty, using it as a token-level indicator to reduce server queries. However, these approaches may still upload redundant or low-value information. This motivates the need for a more refined filtering mechanism that incorporates not only uncertainty but also contextual relevance to minimize unnecessary communication without compromising output quality.

To this end, we propose an energy-aware hybrid inference framework that jointly leverages uncertainty- and importance-aware. Our method filters uploads using two lightweight and complementary signals: (i) epistemic uncertainty derived from temperature perturbations \cite{oh2024uncertainty}, which is about the lack of confidence in the draft token generated by SLM and (ii) token importance measured via attention-weight statistics that reflects the contextual relevance of the token within the sentence structure. A token is uploaded for verification only when it is both uncertain and important—thereby ensuring that cloud resources are used only where it matters.
We identify the phenomenon of attention collapse, where attention distributions become increasingly flat as autoregressive decoding leading to hallucination or factual drift\cite{tang2025seeing}. To address this, we design a dynamic importance threshold that adapts to the distribution of attention weights at each decoding step. It filters out tokens with weak or diffuse attention, preventing unnecessary LLM queries while preserving the token uploads that are considered important.

As shown in Fig.~\ref{Fig:arch}, our framework enables low-overhead, high-fidelity inference by leveraging local SLM statistics to selectively invoke the LLM. For instance, one of our proposed configurations achieves a BERTScore of 87.5\%---nearly matching the HLM's 87.6\%---while reducing energy saving by 40.7\%, lowering the upload rate from 100\% to 38.6\%, and improving token throughput from 0.25 to 0.37. Also, our previous U-HLM~\cite{oh2024uncertainty} performs well, but our method---under a different setting---improves BERTScore (85.8\%→87.0\%), energy savings (31.6\%→43.6\%), and throughput (0.36→0.40). These benefits are achieved without sacrificing output quality. Moreover, the framework supports a tunable trade-off between accuracy and efficiency through two intuitive parameters controlling upload opportunistic, enabling flexible adaptation to system constraints such as latency, bandwidth, and energy.

In summary, this paper makes the following contributions:
\begin{itemize}
    \item We propose a novel uncertainty- and importance-aware speculative decoding framework that opportunistically skips LLM verification based on local token statistics. To mitigate attention collapse, we design an adaptive importance threshold that adjusts dynamically based on the distribution of attention weights at each decoding step.
    \item We provide extensive evaluations showing that our framework significantly reduces LLM usage, bandwidth, and energy costs—while maintaining or exceeding the accuracy of prior methods.
    \item We show that our framework is tunable: the strictness of the upload condition can be adjusted to achieve desired trade-offs across accuracy, latency, and energy efficiency.
\end{itemize}

The remainder of this paper is organized as follows. Section II introduces the system and wireless communication model. Section III presents the proposed opportunistic skipping mechanism based on token uncertainty and importance. Section IV evaluates the performance of our method in terms of accuracy, latency, token throughput, and energy efficiency. Section V concludes with key findings and potential future directions.

\section{System Model}

This section describes the system model, encompassing the hybrid inference architecture and the wireless channel. As described in Fig.~\ref{Fig:arch}, we consider a network with a single user edge device and a base station (BS). The resource-limited edge device, equipped with a SLM, connects to a power-rich BS hosting a LLM over a wireless link. Both entities cooperatively sample with the minimum unit, tokens from a vocabulary $\mathcal{V}$ which contains the full set of possible tokens.

\subsection{Inference Model}
Inspired by our prior hybrid inference framework~\cite{oh2024uncertainty}, the process follows an autoregressive decoding loop, where each step generates a token conditioned on the preceding sequence. At decoding step $t$, the SLM on the edge processes the current sequence $s(t)$, based on the previously generated sequence $s_{1:t-1}$, to compute a probability distribution $\mathbf{x}(t)= [x_1(t), \cdots, x_{|\mathcal{V}|}(t)]^T$ via $\textit{softmax normalization}$ over its logits. A draft token $d(t)\sim\mathbf{x}(t)$ is then sampled and tentatively appended to the sequence, forming a candidate continuation $s(t) = s_{1:t-1} \oplus d(t)$, where $\oplus$ denotes the concatenation operator.

Unlike conventional speculative decoding$-$where all draft tokens and probability distributions are transmitted to server-hosted LLM for verification$-$our opportunistic speculative decoding allows the edge device to locally decide whether the draft token $d(t)$ should be uploaded for verification. If deemed unnecessary, the draft token is accepted locally and added to the response sequence without LLM involvement.
If the draft token $d(t)$ is uploaded, the LLM computes its own distribution $\mathbf{y}(t) = [y_1(t), y_2(t), \cdots, y_\mathcal{|V|}(t)]^T$ in a similar manner. Verification employs the Metropolis-Hastings acceptance criterion \cite{chib1995understanding}, comparing the draft probability $x_d(t)$ with the corresponding LLM probability $y_d(t)$\cite{leviathan2023fast}. If $x_d(t) \leq y_d(t)$, the draft token is immediately accepted; otherwise, it is rejected with probability $1-y_d(t)/x_d(t)$. Upon rejection, the LLM forms a normalized distribution by evaluating the difference between the vocabulary distributions of the LLM and the SLM, assigning the $v$-th probability as:
\begin{equation}
    P_v(t) = \frac{max(y_v(t)-x_v(t), 0)}{\sum_{i=1}^{|\mathcal{V}|}max(y_i(t)-x_i(t), 0)}, \forall{v}\in\mathcal{V},
\end{equation}
and resamples a new target token as $d^*\sim norm(max(y_v(t)-x_v(t), 0))$, where $norm(\cdot)$ denotes the normalization operation that produces a valid probability distribution over $\mathcal{V}$.
The resampled token $d^*(t)$ replaces the rejected draft and is appended to the sequence $s(t) = s_{1:t-1} \oplus d^*(t)$. Each accepted token, whether produced by the SLM or verified/resampled by the LLM, is referred to as a response token in this context.
This process repeats until a maximum length $\vert s(t) \vert=s_{max}$ is reached or an End-of-Sentence (EOS) token is generated.
% This decoding loop continues iteratively until the number of response tokens reaches a predefined maximum length, $\vert s(t) \vert=s_{max}$, or an End-of-Sentence (EOS) token, $s(t)=$EOS, is generated, at which point the inference process terminates.

\subsection{Wireless Channel Model}

We model the edge-server communication as a time-slotted uplink and downlink channel with opportunistic transmission.

\noindent \textbf{Channel Model}. At each decoding step \( t \), the SLM may upload the draft token \( d(t) \) and its associated probability distribution \( \mathbf{x}(t) \) to the server. When the full distribution is uploaded, the uplink payload size is given by $B = |\mathcal{V}| \cdot b_{\mathrm{prob}}$ bits, where $|\mathcal{V}|$ is the vocabulary size, and $b_{\mathrm{prob}}$ denotes the bits precision per probability value.
%(e.g., $b_{\mathrm{prob}}$ = 32 bits for full precision and $b_{\mathrm{prob}}$=16 bits for $float$16 representation). 

The uplink transmission time \( \mu(t) \) is modeled using Shannon's capacity formula under a block fading:
\begin{equation}
\mu(t) = \frac{B}{W \log_2 \left( 1 + \rho(t) \right)},
\end{equation}
where $W$ is the channel bandwidth and $\rho(t)$ is the signal-to-noise ratio (SNR) at time step \( t \). The SNR is defined as $\rho(t) = \frac{h(t) \cdot p \cdot d^{-\alpha}}{\sigma^2}$, where $h(t)$ is the channel gain assumed to be a constant at each time step, $p$ is the transmission power, $d$ is a distance between the edge device and the server, \( \alpha \) is the path loss exponent, and \( \sigma^2 \) is the noise power. 

% b_prob = 32 (B = 32000 * 32 = 1,024,000), rho(t) = 0.196, mu(t) = 3.94 sec // b_prob = 16 (B = 512,000), mu(t) = 1.97 sec
\noindent \textbf{Token Throughput}. The token throughput \( \lambda(t) \) measures the number of tokens generated per unit time, incorporating both communication (uplink only, as downlink overhead is negligible) and computation delays:

\begin{equation}
\lambda(t) (\text{in token/sec}) =
\begin{cases}
\displaystyle\frac{1}{\mu_{\mathrm{SLM}} + \mu(t) + \mu_{\mathrm{LLM}}}, & \text{if } \delta(t) = 1, \\[2ex]
\displaystyle\frac{1}{\mu_{\mathrm{SLM}}}, & \text{if } \delta(t) = 0,
\end{cases}
\end{equation}
where \( \delta(t) \in \{0,1\} \) indicates whether uplink communication and LLM computation are invoked (\( \delta(t) = 1 \)) or skipped (\( \delta(t) = 0 \)). Here, \( \mu_{\mathrm{SLM}} \) and \( \mu_{\mathrm{LLM}} \) represent the inference latencies of the SLM and LLM, respectively.
Specifically, reducing upload frequency (i.e., minimizing the number of cases that $\delta(t)=1$) directly decreases the communication latency $\mu(t)$ and LLM inference time $\mu_{\mathrm{LLM}}$, improving the overall token throughput.
% The latency of local SLM inference \( \mu_{\mathrm{SLM}} \) is much smaller than that of the LLM \( \mu_{\mathrm{LLM}} \) and the uplink transmission delay \( \mu(t) \).
% Therefore, reducing the number of uploads contributes to bandwidth reduction.

However, excessive skipping may degrade inference accuracy, as some tokens may carry high uncertainty or semantic relevance that the SLM cannot handle reliably. Thus, \textit{opportunistic skipping} is introduced: it selectively skips uploads when the SLM’s prediction is likely to be accepted by the LLM, reducing communication costs without compromising generation quality. This trade-off is central to our inference strategy, which balances upload decisions based on local uncertainty and importance evaluations, enabling high-throughput yet accurate text generation.

\section{Opportunistic Skipping Mechanism}

To minimize unnecessary communication and improve end-to-end efficiency in speculative decoding, we introduce a dynamic token selection mechanism termed \textit{opportunistic skipping}. At each decoding step, the system evaluates the draft token $d(t)$ generated by the on-device SLM using two complementary metrics---\textit{token uncertainty} and \textit{token importance}---to decide whether to upload the token to the LLM for verification or commit it locally.

\subsection{Token Uncertainty}
Token uncertainty reflects the SLM’s epistemic confidence in its draft prediction. Drawing inspiration from the uncertainty-aware HLM approach \cite{oh2024uncertainty}, we quantify uncertainty by applying \textit{temperature perturbation} to the SLM's logit outputs. Specifically, for a given decoding step $t$, we generate $N$ perturbed softmax distributions $\{\mathbf{x}^{(i)}(t)\}_{i=1}^{N}$ 
by scaling the logits with randomly sampled temperatures $\zeta_i \in \mathcal{T}$, where $\mathcal{T}$ is a predefined temperature range. From each perturbed distribution $\mathbf{x}^{(i)}(t)$, a token $d_n\sim\mathbf{x}^{(i)}(t)$ is sampled, resulting in a set of tokens $\{{d_n}\}_{i=1}^{N}$. The predicted token probability across these samples serves as the token's uncertainty $u(t)$ for the draft token $d(t)$ as:
% N : temperature_len, 20 / \mathcal{T} : temperature range, np.random.uniform(0, 2, N) / zeta_1 ~ zeta_20
% 근데 zeta를 굳이 정의해야할 필요가 있을까? by scaling the SLM logits with randomly sampled temperatures from a predefined range \mathcal{T}. 로 하면 끝 아님 ??
\begin{equation}
u(t) = \frac{1}{N} \sum_{n=1}^N \mathds {1}(d_n \neq d(t)),
\end{equation}
where $\mathds{1}(d_n \neq d(t))$ returns 1 if the sampled token \( d_n \) differs from the draft token \( d(t) \), and 0 otherwise. A high $u(t)$ indicates that the SLM's prediction is unstable under perturbation, suggesting a higher likelihood of rejection by the LLM.
\begin{figure}[t]	
	\centering	
	\includegraphics[width = 0.45\textwidth]{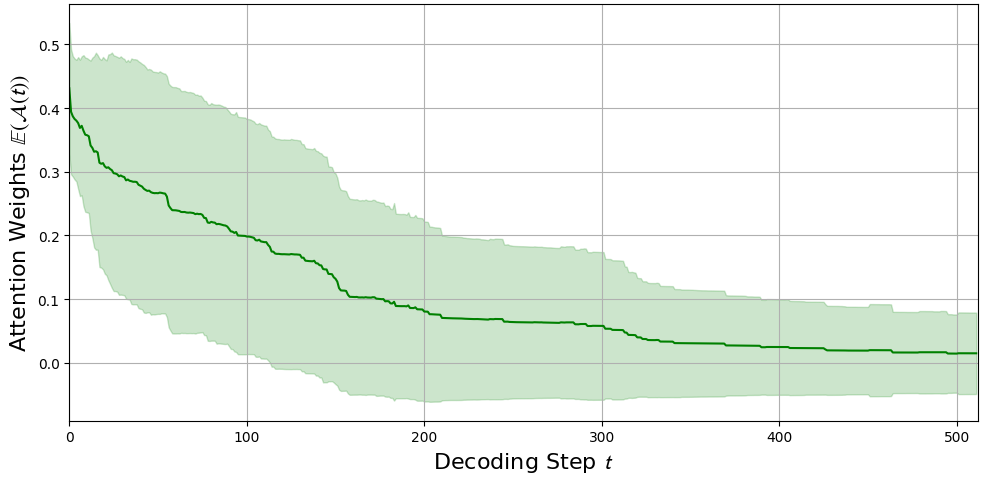}			
	\caption{Attention collapse via decoding step $t$}
	\label{fig:Collapse}
\end{figure}
\subsection{Token Importance}

\begin{table*}[t]
\scriptsize
\centering
\caption{Comparative results across SLM, HLM, U, I, and (U+I) with varying $k$ and $\gamma$.}
\label{tab:overall_table}
\begin{tabular}{|c|c|c|cc|c|c|c|c|}
\hline
\textbf{Method} & \textbf{$k$} & \textbf{$\gamma$} & \multicolumn{2}{c|}{\textbf{Accuracy (\%)}} & \textbf{Upload Rate (\%)} & \textbf{Reject Rate (\%)} & \textbf{Energy Saving Ratio(\%)} & \textbf{Token throughput (tokens/sec)}\\
\cline{4-5}
& & & \textbf{BERT} & \textbf{CS} & & & ($\eta$) & ($\lambda$)\\
\hline
\multirow{1}{*}{SLM} & -- & -- & 85.05 & 44.3 & - & - & 69.5 & 0.53\\
\hline
\multirow{1}{*}{HLM} & -- & -- & 87.6 & 56.7 & 100.0 & 29.11 & - & 0.25\\
\hline
\multirow{1}{*}{U} & -- & -- & 85.8 & 54.5 & 42.06 & 54.58 & 31.6 & 0.36\\
\hline
\multirow{5}{*}{I}
& 3 &  & 86.3 & 51.8 & 70.08 & 29.99 & 17.8 & 0.30\\
& 5 &  & 85.9 & 48.3 & 45.98 & 27.88 & 23.2 & 0.35\\
& 7 & 1.0 & 85.5 & 46.9 & 20.44 & 28.95 & 43.5 & 0.43\\
& 9 &  & 85.5 & 44.0 & 9.04 & 27.34 & 61.7 & 0.48\\
& 11 &  & 85.1 & 44.9 & 4.83 & 28.75 & 60.3 & 0.50\\
\hline
\multirow{15}{*}{U+I} 
& & 0.5 & 86.0 & 50.1 & 13.40   & 47.80 & 57.5 & 0.46\\
& 3 & 1.0 & 87.0 & 54.9 & 28.63 & 58.15 & 43.6 & 0.40\\
& & 1.5 & 87.5 & 56.6 & 38.61 & 57.80 & 40.7 & 0.37\\
\cline{2-9}
&  & 0.5 & 85.7 & 49.2 & 4.27   & 45.28 & 66.3 & 0.50\\
& 5 & 1.0 & 86.4 & 50.8 & 19.61 & 52.73 & 53.5 & 0.43\\
&  & 1.5 & 87.1 & 53.8 & 28.73   & 56.26 & 48.5 & 0.40\\
\cline{2-9}
&  & 0.5 & 84.9 & 47.4 & 0.96   & 31.35 & 69.2 & 0.52\\
& 7 & 1.0 & 86.3 & 48.6 & 14.06 & 46.47 & 59.6 & 0.45\\
&  & 1.5 & 87.0 & 53.5 & 25.78   & 53.39 & 44.3 & 0.41\\
\cline{2-9}
&  & 0.5 & 85.0 & 46.1 & 0.42   & 41.38 & 67.6 & 0.52\\
& 9 & 1.0 & 85.7 & 48.0 & 6.66  & 49.80 & 62.7 & 0.49\\
&  & 1.5 & 86.6 & 52.3 & 17.43   & 52.15 & 52.1 & 0.44\\
\cline{2-9}
&  & 0.5 & 85.1 & 45.0 & 0.08  & 75.00 & 66.0 & 0.53\\
& 11 & 1.0 & 85.5 & 47.4 & 3.27  & 45.89 & 68.9 & 0.51\\
&  & 1.5 & 86.4 & 51.6 & 13.32  & 50.62 & 55.8 & 0.46\\
\hline
\end{tabular}
\end{table*}
% While \textit{uncertainty} captures prediction confidence, it alone is insufficient for guiding upload decisions. A token may be uncertain but contextually unimportant. 
While \textit{uncertainty} reflects prediction confidence, it may still lead to uploading tokens that are contextually unimportant. To address this, we introduce \textit{token importance}, which aims to estimate how contextually relevant a draft token is within its current context \cite{guo2024attention}. %, sun2023simple

This importance is measured based on the token's self-attention pattern, particularly the distribution of attention weights in its attention score vector. At decoding step $t$, let $L(t)$ denote the total length of the input sequence including the prefix and the generated tokens up to $d(t)$, i.e., $L(t) = |s(t)|$. The normalized attention score matrix is denoted as $\mathcal{A}(t) \in \mathbb{R}^{L(t) \times L(t)}$, where each element $a_{i,j}$ represents the attention weight assigned by the $i$-th query vector $\mathbf{q}_i$ to the $j$-th key vector $\mathbf{k}_j$:
\begin{equation}
    a_{i,j} = \frac{\exp\left( \frac{\mathbf{q}_i \cdot \mathbf{k}_j}{\sqrt{d_\mathbf{k}}} \right)}{\sum_{l=1}^{i-1} \exp\left( \frac{\mathbf{q}_i \cdot \mathbf{k}_l}{\sqrt{d_\mathbf{k}}} \right)}, \quad \mathcal{A}(t) = [a_{i,j}]_{1 \leq i,j \leq L(t)}
\end{equation}

\noindent where $d_\mathbf{k}$ is the dimensionality of key/query embeddings. To focus on the attention distribution of the current draft token $d(t)$, we denote by $A_i = [a_{i,1}, \dots, a_{i,i-1}]$ the attention weight vector of the $i$-th row of $\mathcal{A}(t)$. This vector represents the attention from the draft token to all preceding tokens. This quantifies how strongly the draft token attends to prior tokens. 

As decoding progresses, depicted Fig. \ref{fig:Collapse}, attention distributions may become increasingly skewed or flattened---a phenomenon known as \textit{attention collapse}~\cite{tang2025seeing}. In such cases, the model’s attention mass disproportionately concentrates on a few prior tokens, reducing contextual diversity.
% and leading to hallucination or factual drift.
To mitigate this, inspired by \cite{wang2024latte, zhou2022energon}, we adopt the \textit{importance threshold} which is an adaptive thresholding mechanism based on the structural characteristics of the attention vector $A_{i}$. The goal is to selectively upload only those tokens whose attention patterns are statistically grounded and semantically meaningful.

Let $\theta_{\text{imp}}(t)$ denote the importance threshold at step $t$. We define it based on the dynamic range and dispersion of $A_i$ :
\begin{equation}
\label{eqn:imp_threshold}
\theta_{\text{imp}}(t) = \max(A_i) - \tau_{\text{imp}}
\end{equation}
Here, $\theta_{\text{imp}}$ acts as a filtering threshold anchored to the maximum attention score, selectively retaining only high-importance connections. The margin term $\tau_{\text{imp}} = \gamma \cdot \text{std}(A_{i})$, where $\text{std}(A_{i})$ is the standard deviation of the attention weights and $\gamma \in \mathbb{R}^+$ is a tunable scaling factor, modulates the strictness of the filter. Intuitively, $\tau_{\text{imp}}$ represents a statistical gap below $\max(A_{i})$; attention weights smaller than this margin are considered contextually weak or ambiguous and thus filtered out. This thresholding strategy ensures that tokens with sharp, confident attention distributions are evaluated more strictly, while those with flat or noisy distributions---often symptomatic of attention collapse---are less likely to be uploaded.

\subsection{Selective Upload Policy}
To make upload decisions, we define $\tilde{\mathcal{A}}(t)$ as the element-wise ascending sort of all values in $\mathcal{A}(t)$, i.e., $
\tilde{\mathcal{A}}(t) = [a_t^{(1)}, a_t^{(2)}, \dots, a_t^{(L(t)^2)}]$ and $a_t^{(1)}\leq a_t^{(2)}\leq\cdots \leq a_t^{(L(t)^2)}$. $\tilde{\mathcal{A}}(t)$ is obtained by flattening and sorting all elements of $\mathcal{A}(t)$. We then identify the top-$k$ attention weight as $a_t^{(L(t)^2 - k + 1)}$. If $a_t^{(L(t)^2 - k + 1)} > \theta_{\text{imp}}(t)$, the draft token $d(t)$ is considered to have sufficiently strong contextual dependencies and is thus eligible for upload verification.

We combine uncertainty and importance to define a \textit{selective upload policy}:

\begin{equation}
\delta(t) =
\begin{cases}
1, & \text{if } u(t) > \theta_u \text{ and } a_t^{(L(t)^2 - k + 1)} > \theta_{\text{imp}}(t), \\
0, & \text{otherwise},
\end{cases}
\label{eqn:upload}
\end{equation}
where $\theta_u$ is the uncertainty threshold. $\delta(t) = 1$ indicates that the draft token $d(t)$ is uploaded to the LLM for verification; otherwise, it is committed by the SLM.

\begin{figure}[t]
    \centering
    \begin{subfigure}[b]{0.45\textwidth}
        \centering
        \includegraphics[width=0.9\textwidth]{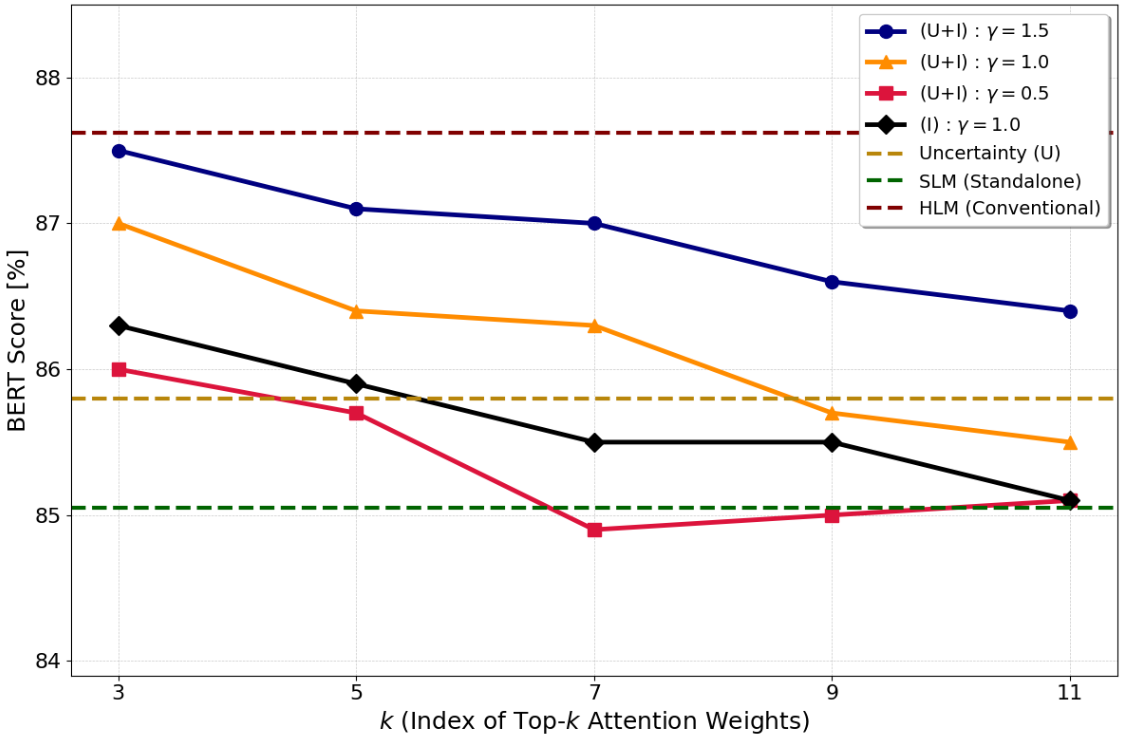}
        \caption{Accuracy of U, I, and U+I according to top-$k$ selection and $\gamma$ based on BERT Score}
        \label{Fig:accuracy_metrics}
    \end{subfigure}
    \hfill
    \begin{subfigure}[b]{0.48\textwidth}
        \centering
        \includegraphics[width=0.9\textwidth]{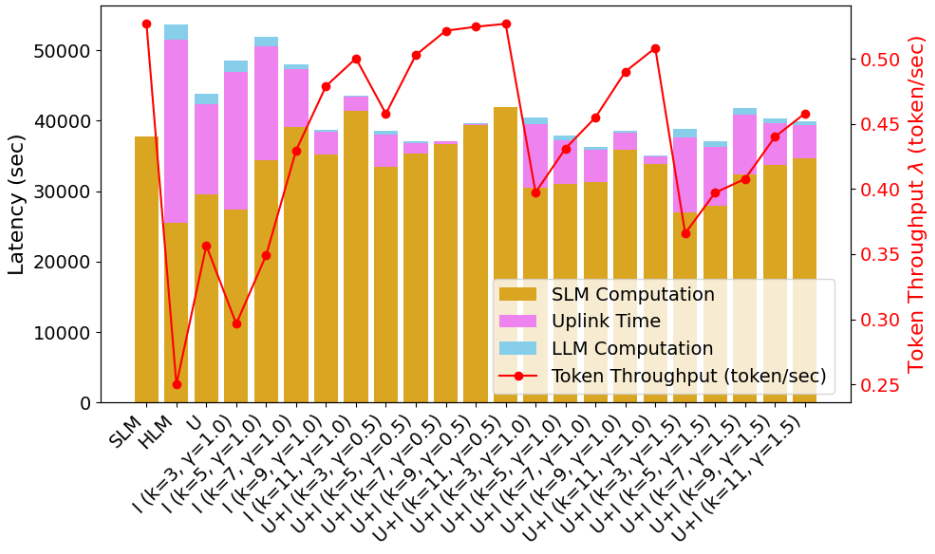}
        \caption{Total inference latency and Token throughput $\lambda$ of U, I, and U+I}
        \label{Fig:Latency}
    \end{subfigure}
    \hfill
    \begin{subfigure}[b]{0.48\textwidth}
        \centering
        \includegraphics[width=0.9\textwidth]{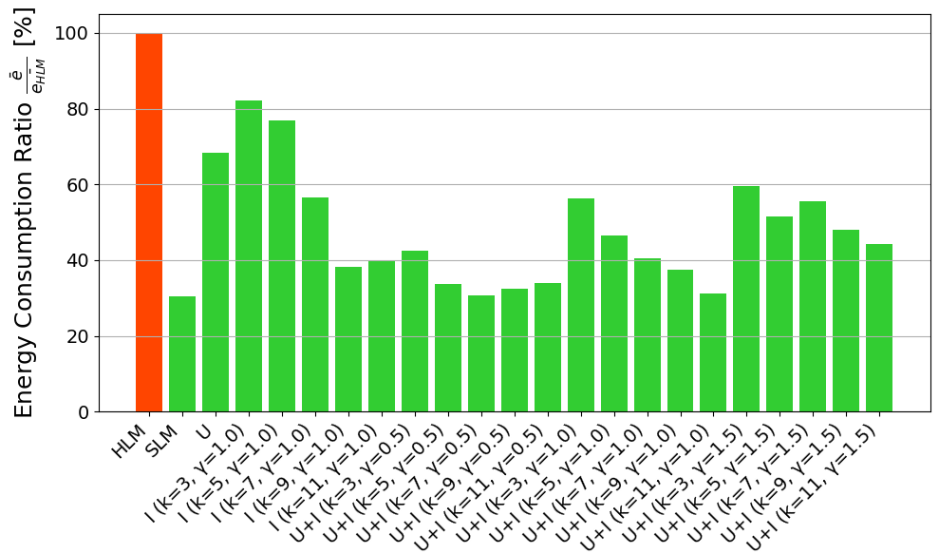}
        \caption{Relative energy consumption (\%) of U, I, and U+I}
        \label{Fig:energy}
    \end{subfigure}
    \caption{Evaluations of U, I, and U+I across various configurations: (a) Accuracy, (b) Latency\&Token throughput, and (c) Energy consumption.}
\end{figure}
\section{Numerical Evaluations}
\noindent\textbf{Experimental Setup}. We conduct our experiments using \texttt{TinyLlama-1.1B} \cite{zhang2024tinyllama} as the on-device small language model (SLM) and \texttt{LLaMA-2-7B} \cite{touvron2023llama} as the cloud-based large language model (LLM). For inference input, we randomly sample 100 prompts from the Alpaca-train-10k dataset \cite{taori2023stanford}, which is widely used for instruction-tuned language generation benchmarks.
All experiments are performed on a Linux-based server equipped with an 8-core Intel Xeon Silver 4215R CPU, 64 GB of DDR4 RAM, and three NVIDIA GeForce RTX 3090 GPUs. The local SLM operates on a single GPU, while the remote LLM is separately invoked for token verification and resampling.

To evaluate the performance of our proposed \textit{Uncertainty-and-Importance-aware HLM (U+I)} framework, we simulate various upload strategies under different configurations:
\begin{itemize}
\item \textbf{SLM (Standalone)}: The SLM generates all tokens without any LLM intervention.
\item \textbf{HLM} \cite{leviathan2023fast}: Every token from the SLM is verified by the LLM, mimicking speculative decoding without filtration.
\item \textbf{Uncertainty (U, U-HLM, U-only)} \cite{oh2024uncertainty}: Tokens are uploaded solely based on epistemic uncertainty $u(t)$.
\item \textbf{Importance (I, I-only)}: Upload occurs only when a token’s attention-based importance exceeds a given threshold $\theta_{\text{imp}}$.
\item \textbf{Uncertainty and Importance (U+I)}: Our proposed method, which uploads a token only if both uncertainty and importance conditions are simultaneously satisfied.
\end{itemize}
\noindent For the importance condition, we explore two configurable parameters: \textbf{Top-$k$ selection} and \textbf{$\gamma$ scaling}. A token is deemed important if at least $k$ of its attention weights exceeding the importance threshold $\theta_{\text{imp}}(t)$, as introduced in Section III-C. Larger values of $k$ enforce stricter filtering. We evaluate $k \in \{3, 5, 7, 9, 11\}$, omitting $k=1$ due to its triviality. 

\noindent The $\gamma$ scaling factor governs the margin $\tau_{\text{imp}}$ in Eq.~\eqref{eqn:imp_threshold}, defined as $\tau_{\text{imp}} = \gamma \cdot \mathrm{std}(A_t)$. A higher $\gamma$ loosens the filtering condition by lowering $\theta_{\text{imp}}$, allowing more tokens to pass—potentially improving accuracy while increasing LLM usage. We test with $\gamma \in \{0.5, 1.0, 1.5\}$.
% \begin{figure}[h]	
% 	\centering	
% 	\includegraphics[width = 0.45\textwidth]{Figure/BERT score6.png}			
% 	\caption{Accuracy of U, I, and U+I according to top-$k$ selection and $\gamma$ based on BERT Score}
% 	\label{Fig:accuracy_metrics}
% \end{figure}

Channel settings follow practical edge-to-cloud assumptions: The vocabulary size $|\mathcal{V}| = 32,000$, the bit precision $b_{\mathrm{prob}} = 32$ or $16$ $\text{bits}$, so the uplink payload size $B$ is computed as \( B = |\mathcal{V}| \cdot b_{\mathrm{prob}} \). The uplink channel bandwidth $W = 1 \, \text{MHz}$, the transmission power $p=23$ $\text{dBm}$, noise $\sigma^2=-104$ $\text{dBm}$, the distance between the edge device and the server $d = 2.5$ $\text{km}$, and the path loss exponent
$\alpha = 4$. Rayleigh fading is modeled as $h(t) \sim \exp(1)$, with average-case $h(t) = 1$.

We evaluate the results with respect to accuracy, latency, token throughput, and energy efficiency. These are summarized in detail in the TABLE \ref{tab:overall_table}.

\noindent \textbf{Accuracy.} As shown in Fig.\ref{Fig:accuracy_metrics}, we analyze accuracy in terms of BERT Score (BERT)\cite{zhang2019bertscore} and Cosine similarity (CS). U+I demonstrates superior accuracy across multiple configurations. U+I with ($k$=3, $\gamma$=1.5) achieves a BERT of 87.5\%, closely matching HLM (87.6\%), while significantly reducing LLM queries. Similarly, it shows higher CS of 55.7\%, closely rivaling HLM (56.7\%).
% Similarly, it shows higher CS (55.7\%) than HLM (56.7\%) with comparable reliability. 
Across all $\gamma$, the accuracy generally decreases as the top-$k$ parameter increases, due to fewer tokens being uploaded for verification. However, U+I gracefully handles this trade-off, and retains robust performance even with low upload rates causing energy savings.
% \begin{figure}[h]	
% 	\centering	
% 	\includegraphics[width = 0.49\textwidth]{Figure/Latency&throughput4.png}			
% 	\caption{Total inference latency and Token throughput $\lambda$ of U, I, and U+I}
% 	\label{Fig:Latency}
% \end{figure}

%% Latency 수정 중!
\noindent \textbf{Latency \& Token Throughput}. As depicted in Fig.\ref{Fig:Latency}, we compares the latency, which is measured as the total inference time for a fixed-size instruction set, and token throughput that denotes the number of generated tokens per second. HLM achieves the highest accuracy but suffers from the longest latency and the lowest throughput of 0.25 due to full LLM verification. In contrast, SLM performs all decoding locally without server involvement, resulting in the highest throughput of 0.53 and very low latency, but at the cost of degraded quality of 85.05\% BERT. U-only and I-only offers intermediate performance. Both accomplish throughput of 0.36 and 0.30-0.50 along with LLM interactions, respectively, while reducing latencies compared to HLM. However, they need to compromise with accuracy. 
U+I leverages its dual-condition to obtain notable reductions in latency and improvements in token throughput while preserving quality. For example, ($k$=3, $\gamma$=1.5) achieves 87.5\% BERT, while improving throughput to 0.37 tokens/sec. Other configurations such as ($k$=5, $\gamma$=1.0) and ($k$=7, $\gamma$=1.5) also strike favorable trade-offs, attaining 86.4-87.0\% BERT with throughput exceeding 0.40 tokens/sec.

As $k$ increases or $\gamma$ decreases, upload conditions become stricter, leading to reduced LLM usage and improved throughput. For instance, ($k$=11,$\gamma$=0.5) invokes the LLM for only 0.08\% of tokens while maintaining high throughput, but at a slight accuracy drop of 85.1\% BERT. Conversely, smaller $k$ and higher $\gamma$ yield more liberal uploads, increasing LLM reliance and reducing throughput but boosting output quality.

% achieve, accomplish, fulfill, attain, obtain, gain, get

% In contrast, U+I leverages its dual-condition mechanism to avoid unnecessary uploads while preserving quality. Its configurable ($k$, $\gamma$) parameters enable adaptation to system constraints. Lower $k$ or higher $\gamma$ values relax upload conditions, leading to higher throughput and accuracy, but increase latency due to more LLM use. Conversely, increasing $k$ or reducing $\gamma$ results in stricter filtering, reducing latency but potentially sacrificing accuracy.

% \begin{figure}[h]	
% 	\centering	
% 	\includegraphics[width = 0.49\textwidth]{Figure/Energy saving2.png}			
% 	\caption{Relative energy consumption (\%) of U, I, and U+I}
% 	\label{Fig:energy}
% \end{figure}

\noindent \textbf{Energy Efficiency}. Inspired by \cite{cao2018joint}, we evaluate the total energy consumption as $\bar{e}=e_u+e_r+e_s$, where uplink transmission energy $e_u=l_u \cdot B\cdot\epsilon_u$, LLM resampling energy $e_r=l_r \cdot \epsilon_r$, and SLM inference energy $e_s=l_s \cdot \epsilon_s$. Here, $l_u,l_r,$ and $l_s$ denote the number of uploaded, rejected, and locally processed tokens, respectively; and $\epsilon_u, \epsilon_r, \epsilon_s$ are the per-token power costs set empirically to \{300, 350, 100\}\text{W}. To quantify relative efficiency, we define the energy saving ratio as $\eta=1-\frac{\bar{e}} {\bar{e}_{\text{HLM}}}$, where $\bar{e}_{\text{HLM}}$ is the baseline energy consumption of HLM. 
The $\eta$ captures how much energy is saved relative to the full inference used in HLM.
% The $\eta$ represents the relative reduction in energy use compared to the full inference used in HLM.

As shown in Fig~\ref{Fig:energy} and Table~\ref{tab:overall_table}, HLM exhibits the highest energy usage since it uploads and verifies every token, while SLM saves the most energy $\eta_{SLM}=69.5\%$ by avoiding any LLM usage, albeit with reduced accuracy 85.05\% BERT.
U-only offers moderate efficiency $\eta_u=31.6\%$ by reducing the upload rate to 42.06\%, though it incurs a high reject rate of 54.58\%, which amplifies resampling overhead. I-only achieves higher efficiency with stricter upload policies; for instance, $k=9$ yields 61.7\% energy reduction, though at the cost of lower LLM involvement and possible degradation in accuracy.
Our proposed U+I balances uncertainty and importance to enable tunable LLM usage via $k$ and $\gamma$, providing substantial energy savings across a wide range of settings. As $k$ increases or $\gamma$ decreases, upload conditions tighten, reducing LLM usage and improving throughput—but potentially lowering accuracy. In more aggressive cases like ($k$=11,$\gamma$=0.5), it reduces LLM involvement to near-zero, achieves 66.0\% savings, at a small cost in accuracy of 85.1\%.
% High $k$ combined with low $\gamma$ leads to a stricter upload condition, resulting in minimal LLM usage and reduced latency, thus enhancing energy efficiency—but at the cost of lower accuracy. 
Conversely, low $k$ and high $\gamma$ relax the filtering, allowing more tokens to be uploaded, which improves accuracy but increases energy consumption and latency. U+I ($k$=3, $\gamma$=1.5) achieves near-HLM accuracy with 40.7\% energy savings.
Intermediate configurations such as ($k$=5, $\gamma$=1.0) or ($k$=7, $\gamma$=1.0) may strike the best balance between performance and efficiency. U+I ($k$=5, $\gamma$=1.0) offers a compelling trade-off, saving 53.5\% energy while preserving accuracy 86.4\% BERT, and U+I ($k$=9, $\gamma$=1.0) fulfill saving 59.6\% energy while sustaining a 86.3\% BERT.

Through this tunability, U+I dynamically adapts to both semantic relevance and epistemic uncertainty
% —unlike static baselines like HLM or U-only—
and enables fine-grained control over the efficiency-quality balance.

\section{Conclusion}
In this paper, we presented an importance- and uncertainty-aware speculative decoding framework for efficient HLM inference. Our method substantially reduces LLM usage while preserving or even improving accuracy. Extensive experiments demonstrate that tunable parameters $k$ and $\gamma$ allow for adaptive trade-offs between accuracy, latency, and energy. This makes our approach practical for deployment in bandwidth-constrained or energy-sensitive edge environments.
Future work may extend our approach to multi-access settings, where multiple SLMs collaboratively interact with a shared LLM server.

\section{Acknowledgements}
This work was supported by Institute of Information \& communications Technology Planning \& Evaluation(IITP) grant funded by the Korea government(MSIT) (No.2022-0-00420,Development of Core Technologies enabling 6G End-to-End On-Time Networking)

\bibliographystyle{ieeetr}
\bibliography{./ref.bib}

\end{document}